\documentclass[pdflatex,iicol]{sn-jnl}% Math and Physical Sciences Numbered Reference Style
\usepackage{graphicx}%
\usepackage{multirow}%
\usepackage{amsmath,amssymb,amsfonts}%
\usepackage{amsthm}%
\usepackage{mathrsfs}%
\usepackage[title]{appendix}%
\usepackage{xcolor}%
\definecolor{deepgreen}{RGB}{55,126,34}
\usepackage{textcomp}%
\usepackage{manyfoot}%
\usepackage{booktabs}%
\usepackage{algorithm}%
\usepackage{algorithmicx}%
\usepackage{algpseudocode}%
\usepackage{listings}%
\usepackage{natbib}%
\usepackage{orcidlink}
\usepackage{mathtools}  % Put this in your preamble
\usepackage{subcaption}
\usepackage{tabularx}

\theoremstyle{thmstyleone}%
\theoremstyle{thmstyletwo}%

\theoremstyle{thmstylethree}%

\begin{document}

\title[Article Title]{DUDE: Diffusion-Based Unsupervised Cross-Domain Image Retrieval}


\author[1]{\fnm{Ruohong} \sur{Yang}}\email{ruohong.yrh@gmail.com}

\author[1]{\fnm{Peng} \sur{Hu}}\email{penghu.ml@gmail.com}
\author*[1]{\fnm{Yunfan} \sur{Li}}\email{yunfanli.gm@gmail.com}
% \equalcont{These authors contributed equally to this work and are co-corresponding authors.}
\author*[1]{\fnm{Xi} \sur{Peng}}\email{pengx.gm@gmail.com}
% \equalcont{These authors contributed equally to this work and are co-corresponding authors.}

\affil*[1]{\orgdiv{College
of Computer Science}, \orgname{Sichuan University}, \orgaddress{\city{Chengdu}, \postcode{610065}, \country{China}}}


\abstract{Unsupervised cross-domain image retrieval (UCIR) aims to retrieve images of the same category across diverse domains without relying on annotations. Existing UCIR methods, which align cross-domain features for the entire image, often struggle with the domain gap, as the object features critical for retrieval are frequently entangled with domain-specific styles. To address this challenge, we propose DUDE, a novel UCIR method building upon feature disentanglement. In brief, DUDE leverages a text-to-image generative model to disentangle object features from domain-specific styles, thus facilitating semantical image retrieval. To further achieve reliable alignment of the disentangled object features, DUDE aligns mutual neighbors from within domains to across domains in a progressive manner. Extensive experiments demonstrate that DUDE achieves state-of-the-art performance across three benchmark datasets over 13 domains. The code will be released.}




\maketitle

\section{Introduction}\label{sec1}

\begin{figure*}[t]
  \centering
   \includegraphics[width=\linewidth]{ 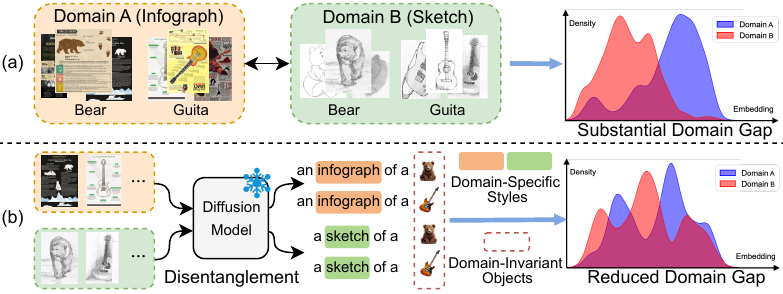}
   \caption{Our key idea. (a) Existing UCIR methods align cross-domain features for the entire image, in which domain-specific styles (\textit{e.g.},  infograph texts and sketch lines) often overshadow the object features, leading to a substantial domain gap. (b) Our method mitigates this gap by disentangling domain-invariant object features from domain-specific styles. In the figure, the right part illustrates the feature density distributions across two domains, before and after disentanglement. 
   }
   \label{fig:motivate}
\end{figure*}

Cross-domain image retrieval (CIR) aims to retrieve relevant images from one domain given a query from another domain \cite{cir3}. CIR has widespread applications, such as matching user-uploaded photos with product listings on shopping platforms \cite{shop1,shop2} and matching forensic sketches with suspect databases in law enforcement \cite{law1,law2}. To effectively address the CIR problem, existing works \cite{cir1,cir2,cir3,cir4} typically rely on annotated labels or paired data as supervision signals, which are costly and labor-intensive to obtain in practice. To reduce the labeling burden, recent efforts have pivoted toward unsupervised cross-domain image retrieval (UCIR) \cite{ucir1,cds,pcs,id}, liberated from manual annotations. To achieve UCIR, most existing methods resort to instance- and cluster-wise contrastive learning for cross-domain alignment \cite{dd,protoot,coda}. A recent study proposes phase-enhanced contrastive learning \cite{hu&small1}, which narrows the domain gap by converting RGB images into phase images.

Despite notable progress, existing UCIR methods still suffer from the domain gap inherent in cross-domain data. This challenge arises from the entanglement of domain-specific styles and domain-invariant objects in images. For instance, as illustrated in Fig.~\ref{fig:motivate} (a), ``an infograph of a bear" contains dense textual descriptions, whereas ``a sketch of a bear" is rendered in simple lines. As existing methods align cross-domain features for the entire image, the prevalence of domain-specific styles can overwhelm the semantically important objects, leading to performance degradation. To enable effective UCIR, it is highly expected that the domain-invariant object features could be disentangled from domain-specific styles. 

However, disentangling the semantic and style features is particularly challenging in the unsupervised setting \cite{disen}. As a classic solution, $\beta$-VAE \cite{beta} suggests that unsupervised feature disentanglement can be achieved based on assumptions of prior distributions. Instead of relying on carefully designed priors, we leverage a text-to-image model as a disentangler. The intuition nestles in the following: if a text-to-image model can synthesize images that faithfully reflect both the style and object specified in a prompt, it should likewise be capable of disentangling style and object from an image in the token space. The disentangled object features offer clean semantic alignment across domains. Compared with exhaustively mitigating the domain gap from entangled image features, it would yield twice the effect with half the effort by disentangling object features from domain-specific styles before the cross-domain alignment.

Based on such a motivation, we propose DUDE, a \textbf{D}iffusion-based \textbf{U}nsupervised cross-\textbf{D}omain image r\textbf{E}trieval method. In brief, DUDE first employs an \textbf{Object Disentanglement Module} to extract domain-invariant object features, leveraging style-specific prompts to counteract domain styles. Specifically, DUDE formulates textual prompts using a handcrafted domain-style token (\textit{e.g.}, painting, infograph, sketch) and a learnable object-semantic token. By deducing the textual prompt that reproduces the same image semantics, DUDE ensures that the object semantic token learns only the clean object features present in the image, given that the image style is encapsulated by the domain-style token.
As illustrated in Fig.~\ref{fig:motivate}(b), the disentangled object features effectively narrow the domain gap. To further enhance cross-domain alignment of these object features, DUDE incorporates a \textbf{Progressive Alignment Module}. By progressively constructing positive contrastive pairs from single instances, in-domain mutual neighbors, to cross-domain mutual neighbors, DUDE achieves reliable and accurate UCIR. Thanks to the proposed disentangle-then-align paradigm, DUDE achieves state-of-the-art performance across 13 domains on three widely used benchmarks.

The main contributions of this work could be summarized as follows:
\begin{itemize}
\item To tackle the core domain gap challenge in UCIR, we propose disentangling domain-invariant object features from domain-specific styles. To the best of our knowledge, this could be the first attempt at solving UCIR from the feature disentanglement perspective.
\item To further align the disentangled object features, we introduce a progressive alignment strategy that gradually expands the positive contrastive pairs, from single instances, within-domain mutual neighbors, to cross-domain mutual neighbors, leading to more accurate UCIR results.
\item The proposed DUDE outperforms existing UCIR methods by a substantial margin on three standard benchmarks over 13 domains. For instance, DUDE achieves an average P@200 improvement of 18.78$\%$ compared with the best competitor on DomainNet.
\end{itemize}

\section{Related Work}\label{sec2}
In this section, we briefly review two fields related to this work, namely, cross-domain image retrieval (CIR) and stylized image generation.

\subsection{Cross-Domain Image Retrieval} 
CIR plays a pivotal role in computer vision across many real-world scenarios, where the query and database images originate from different domains. For example, users may submit freehand sketches as queries to retrieve photos from a database of real photographs with the same category \cite{user,user2,user3}. Although CIR methods have demonstrated strong performance, they heavily rely on large-scale annotations that are labor-intensive to obtain \cite{cir5,cir6,cir1,cir2,cir7}. Such a limitation has aroused several recent efforts in unsupervised cross-domain image retrieval (UCIR) \cite{ucir1,pcs,coda,dd,protoot}. For instance, DD combines intra-domain representation learning with cross-domain alignment to reduce the domain gap \cite{dd}. ProtoOT leverages optimal transport to capture intra-domain consistency and cross-domain correspondence simultaneously \cite{protoot}. Another branch of study adjusts spatial channel-wise or adjusts frequency components of images to narrow the domain gap \cite{small2,small3,small4}. For example, a recent work finds that the domain gap could be reduced in phase images, and thereby designs phase-enhanced contrastive learning to facilitate UCIR \cite{hu&small1}.

Distinct from existing UCIR studies that focus on designing new feature extraction and alignment strategies for the entire image, this work attempts to solve UCIR from a novel disentanglement perspective. Namely, we propose to extract retrieval-relevant object features by disentangling them from domain-specific styles. By extracting clean object semantics before performing cross-domain alignment, our method significantly enhances the retrieval performance.

\subsection{Stylized Image Generation} 
Stylized image generation, a task that involves generating images of the given specific style \cite{style}, has been widely adopted across diverse applications, including creative design \cite{design1,design2}, digital art \cite{art1,art2}, and short video entertainment \cite{video1,video2}. To control the style in image generation, existing methods resort to style-specific fine-tuning \cite{ft1,ft2}, attribute control via conditional variational autoencoders \cite{ac1,ac2}, and prompt-based techniques \cite{pr1,pr2}. For text-to-image generation, style control is commonly achieved by fine-tuning the U-Net on target-style images, as in Stable Diffusion \cite{sd}, or through lightweight fine-tuning methods such as low-rank adaptation \cite{lora}, which injects trainable rank-decomposition matrices for efficient model adaptation.

In addition to generation, diffusion generative models have also demonstrated potential in representation learning \cite{drepre,drepre2,drepre3}, as well as discriminative tasks such as classification \cite{class} and segmentation \cite{seg1,seg2}. Beyond existing applications, this work exhibits diffusion models' capability in cross-domain image retrieval, exploring and mining their feature disentanglement power. We hope this work could inspire further research towards the largely untapped potential of diffusion models for tasks beyond generation.

\section{Methods}\label{sec3}
\begin{figure*}[t]
  \centering
   \includegraphics[width=\linewidth]{ 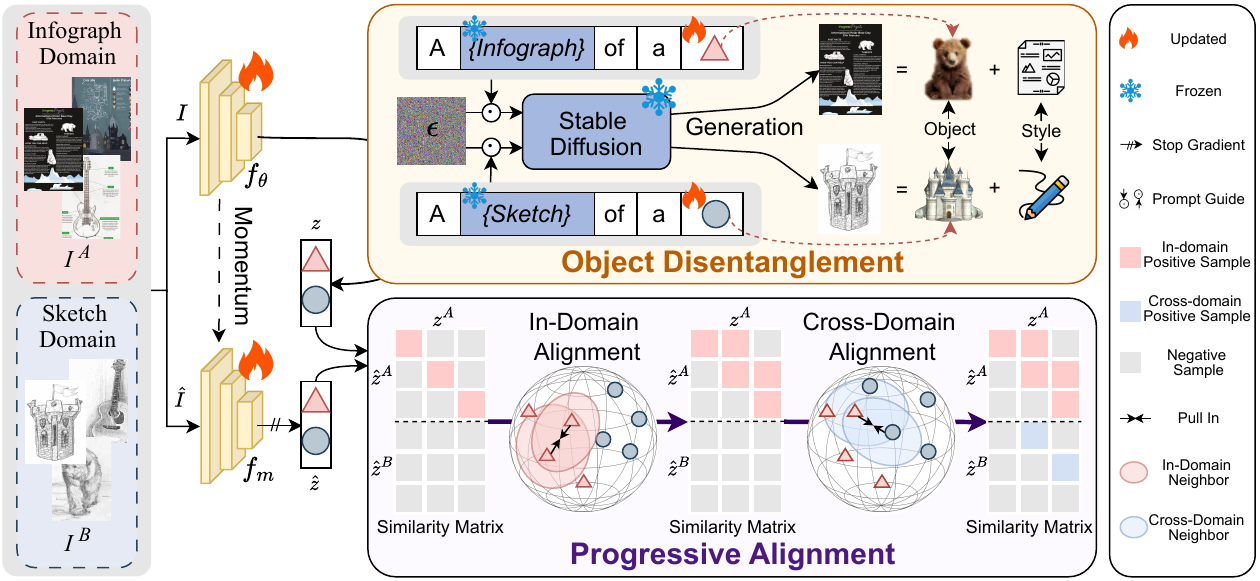}
   \caption{An overview of the proposed DUDE, which consists of two modules:
(a) The \textbf{Object Disentanglement Module} leverages a frozen  Stable Diffusion and varying prompts with different domain-specific tokens (\textit{e.g.}, $\{infograph\}$ and $\{sketch\}$). It guides the feature extractor $f_\theta$ to extract object features by disentangling them from domain-specific styles.
(b) To achieve reliable cross-domain alignment, the \textbf{Progressive Alignment Module} encourages the clean object features $z$ to align with their augmentations, then with in-domain mutual neighbors, and progressively with cross-domain mutual neighbors.}
   \label{fig:method}
\end{figure*}
In this section, we elaborate on the proposed method, DUDE, as illustrated in Fig.~\ref{fig:method}. In brief, Sec.~\ref {sec3.1} provides a brief overview of the preliminaries on unsupervised cross-domain image retrieval (UCIR) and diffusion models. Then, Sec.~\ref{sec3.2} introduces the Object Disentanglement Module built upon Stable Diffusion (SD) to disentangle object features from domain-specific styles. Next, Sec.~\ref{sec3.3} presents a Progressive Alignment Module that constructs positive contrastive pairs from instances, then in-domain neighbors, and gradually cross-domain neighbors, to achieve stable and accurate cross-domain alignment.

\subsection{Preliminary}\label{sec3.1}
We begin with the problem formulation for the UCIR task and a brief overview of SD.

\textbf{Problem Formulation.}
In the UCIR task, we deal with unlabeled images from two domains $D^A=\{I_i^A\}_{i=1}^n, D^B=\{I_i^B\}_{i=1}^m$. Given a query image $I_i^A$ belonging to category $C$, the goal of UCIR is to retrieve images of the same category $C$ from domain $B$. To achieve this, it is required to train a valid feature extractor $f_{\theta}: I_i \rightarrow z_i$ which projects the input image to a common embedding space.

\textbf{Stable Diffusion.}
SD incorporates a Variational Autoencoder (VAE) \cite{vae} $\gamma_{\theta}$ to first encode an image $x \in \mathbb{R}^{512 \times 512}$ into a latent representation $x_0 \in \mathbb{R}^{64 \times 64}$, and then reconstruct an image from the latent representation. To enable high-quality reconstruction, SD employs a denoising process where Gaussian noise is progressively added to the latent representation during the forward process and then removed step by step across $T$ timesteps during the reverse process.

Formally, at timestep $t \in \{1, 2, \dots, T\}$, the noisy latent $x_t$ is computed as:
\begin{equation}
    x_t = \sqrt{\bar{\alpha}_t} \, x_0 + \sqrt{1 - \bar{\alpha}_t} \, \epsilon,
\end{equation}
where $x_0$ is the latent representation, $\bar{\alpha}_t$ is a pre-defined noise schedule parameter, and $\epsilon$ is Gaussian noise sampled from $\mathcal{N}(0, I)$.

To denoise $x_t$ at this timestep, the model is trained to predict the added noise using the following objective:
\begin{equation}\min_\phi \, \mathbb{E}_{\epsilon, x, c, t} \left[ \left\| \epsilon - \epsilon_\phi(x_t, c, t) \right\|_2^2 \right],\end{equation}
where $c$ is the encoded text prompt, and $\epsilon_\phi$ denotes the noise predictor parameterized by $\phi$.

\subsection{Object Disentanglement}\label{sec3.2}

For the UCIR problem, directly extracting features from the entire image often leads to entanglement of domain-invariant objects with domain-specific styles such as sketch lines, textual elements, or richly colored patterns, which degrades cross-domain retrieval performance. Therefore, a critical step is to disentangle domain-invariant objects from these domain-specific styles. 

Recalling that SD can generate consistent objects under different stylistic prompts, this capacity suggests an implicit separation between objects and styles within the prompt token space. Inspired by this, we propose a diffusion-based object disentanglement strategy, leveraging the stylized image generation capabilities of SD to disentangle object features from domain styles.

Let $f_{\theta}: I_i \rightarrow z_i$ be a feature extractor that projects an image $I_i$ into the token space. For each image $I_i^d$ from domain $D^d$ ($d \in \{A, B\}$), we construct a prompt: ``a $\{domain\}$ of a [$z_i^d$]."
Here, $\{domain\}$ is the domain-style token (\textit{e.g.}, painting, infograph, sketch) shared among all samples in domain $D^d$, while the object-semantic token [$z_i^d$] encodes the feature $z_i^d = f_{\theta}(I_i^d)$. This prompt is used to guide the pre-trained SD in reconstructing the image from a noisy latent version of $I_i^d$. Rather than optimizing the image generation capability of SD itself, we freeze all parameters of SD, including the noise predictor $\epsilon_{\phi}$ and the VAE encoder $\gamma_{\phi}$, and train the feature extractor $f_{\theta}$. The training objective is to compress the domain-invariant object feature in $I_i^d$ into the object-semantic token [$z_i^d$].

Formally, given an image $I_i^d$, we first encode $I_i^d$ into a latent representation using the frozen VAE $\gamma_{\phi}$ and calculate the noisy latent at a diffusion timestep $t$ as:
\begin{equation}
    x_{i,t}^d = \sqrt{\bar{\alpha}_t} \, \gamma_{\phi}(I_i^d) + \sqrt{1 - \bar{\alpha}_t} \, \epsilon.
\end{equation}
Then, we compute the disentanglement loss as:
 \begin{equation}
\mathcal{L}_{\text{OD}} = \mathbb{E}_{\epsilon, t} \left[ \left\| \epsilon - \epsilon_{\phi}(x_{i,t}^d, c_{\theta}, t) \right\|_2^2 \right],
\end{equation}
where $c_{\theta}$ is the prompt embedding constructed from the frozen text encoder using the feature $z_i^d$ inserted into the prompt ``a $\{domain\}$ of a [$z_i^d$]", and $\epsilon_{\phi}$ denotes the frozen pre-trained SD noise predictor.

Such a design inverts the typical use of generative models. Rather than generating from text, SD is leveraged as a disentangler of domain-invariant object features [$z_i^d$] and domain-specific styles $\{domain\}$, as gradients are backpropagated to solely optimize $f_{\theta}$.

\subsection{Progressive Alignment}\label{sec3.3}
Thanks to the Object Disentanglement Module, the feature extractor $f_{\theta}$ effectively captures object features while stripping away domain-specific styles. These clean object features, when properly aligned, are expected to form compact clusters for the same class, regardless of their domain of origin. To facilitate reliable and effective alignment, we propose a progressive alignment strategy that constructs positive contrastive samples in stages, from instance to in-domain mutual nearest neighbors, and then to cross-domain mutual nearest neighbors.

Firstly, contrastive learning is conducted between each image and its augmentation, denoted as $\{I_i^d, \hat{I}_i^d\} \in D^d$, where $d \in \{A, B\}$ indicates the domain index. The clean object features are computed as $z_i^d = f_{\theta}(I_i^d)$ and ${\hat{z}_i^d} = f_{m}(\hat{I}_i^{d})$, using the feature extractor $f_{\theta}$ and the momentum extractor $f_m$, respectively. The momentum extractor is updated following the MoCo strategy \cite{moco}, and its outputs are stored in domain-specific memory banks $\mathcal{Q}^d$.

\textbf{In-domain Alignment.}
The most reliable alignment occurs when only augmented views of the same sample are treated as positive pairs, while all other samples in the dataset are considered negatives. The loss function is defined as: 
\begin{equation}
    \mathcal{L}_{\mathrm{aug}}=\sum_{i \in \mathbf{I}}-\log \frac{\exp \left({z}_i^{\top} \hat{z}_i / \tau\right)}{\sum_{a \in \mathbf{I}} \exp \left({z}_i^{\top} \hat{z}_a / \tau\right)},
\end{equation}
where the index set $\mathbf{I}$ represents all the samples in the same domain, and $\tau$ is the temperature parameter.

Next, we extend the alignment progressively to in-domain neighbors. To capture feature relationships within a domain, we define a directed neighborhood graph for each sample $z_i^d$ based on the top-$k$ cosine similarity:
\begin{equation}
\mathcal{N}_k^d(i) = 
\operatorname{Top}\text{-}k\left(\left\{j \mid \cos \left(z_i^d, z_j^d\right)\right\}\right).
\end{equation}

To ensure mutual reliability between neighbors, we define a symmetric adjacency matrix $\mathbf{M}_{\mathrm{in}}^d \in\{0,1\}^{n^d \times n^d}$, where $n^d = |D^d|$. Then the definition becomes:
\begin{equation}
[\mathbf{M}_{\mathrm{in}}^d]_{ij} =
\begin{cases}
1, & \text{if } i \in \mathcal{N}_k^d(j) \text{ and } j \in \mathcal{N}_k^d(i), \\
0, & \text{otherwise.}
\end{cases}
\end{equation}
This adjacency matrix captures only those sample pairs that mutually recognize each other as top-$k$ neighbors, filtering noisy or incidental neighbors, thereby promoting more reliable alignment.

The in-domain alignment loss is formulated using an InfoNCE \cite{infonce} objective, where only mutual neighbors in $\mathbf{M}_{\mathrm{in}}^d$ contribute as positives:
\begin{equation}
\mathcal{L}_{\mathrm{in}}^d = - \frac{1}{{n^d}} \sum_{i=1}^{n^d} \frac{\sum_{j=1}^{n^d} ([\mathbf{M}_{\mathrm{in}}^d]_{ij} \cdot \ell_{ij}^d)}{\sum_{j=1}^{n^d} [\mathbf{M}_{\mathrm{in}}^d]_{ij} + \epsilon},
\end{equation}
\begin{equation}
\ell_{ij}^d = \frac{(z_i^{d}) ^{\top} \hat{z}_j^{d}}{\tau} - \log \sum_{q=1}^{|\mathcal{Q}^d|} \exp \left(\frac{(z_i^{d}) ^{\top} \hat{z}_q^{d}}{\tau}\right),
\end{equation}
where $\hat{z}_j^{d}$ and $\hat{z}_q^{d}$ represent the features in memory bank, $\tau$ denotes the temperature parameter, and $\epsilon$ denotes a small constant to avoid division by zero. By combining the in-domain losses from both domains, the total in-domain loss is defined as: \begin{equation}\mathcal{L}_{\mathrm{in}} = \mathcal{L}_{\mathrm{in}}^A+\mathcal{L}_{\mathrm{in}}^B.
\end{equation}
The progressive loss in the first stage is defined as:
\begin{equation}\mathcal{L}_{\mathrm{PA1}} = \mathcal{L}_{\mathrm{aug}}+\beta\mathcal{L}_{\mathrm{in}}.
\label{eq:PA1}
\end{equation}
% where $\beta$ is set to 0.5 for all datasets.

\textbf{Cross-domain Alignment.}
Once strong in-domain alignment is established, we extend the alignment to cross-domain neighbors. Here, we only describe the alignment process from domain $A$ to domain $B$, as the process from domain $B$ to domain $A$ is symmetric. For each feature $z_i$ extracted from image $I_i^A \in D^A$, we identify its top-$k$ nearest neighbors in domain $B$:
\begin{equation}
\mathcal{N}_k^{A \rightarrow B}(i) = \operatorname{Top}\text{-}k\left(\left\{j \mid \cos \left(z_i^A, z_j^B\right)\right\}\right).
\end{equation}

To reduce noisy neighbors, we further enforce a mutual cross-domain adjacency matrix $\mathbf{M}_{\mathrm{cross}}^{A \rightarrow B} \in \{0,1\}^{n \times m}$:
\begin{equation}
[\mathbf{M}_{\mathrm{cross}}^{A \rightarrow B}]_{ij} =
\begin{cases}
1, & \begin{aligned}[t]
       &\text{if } i \in \mathcal{N}_k^{B \rightarrow A}(j) \\
       &\text{and } j \in \mathcal{N}_k^{A \rightarrow B}(i),
    \end{aligned} \\
0, & \text{otherwise.}
\end{cases}
\end{equation}
This matrix captures only the mutually nearest neighbor pairs across domains, ensuring that alignment occurs only between features with high similarity. The cross-domain loss from domain $A$ to domain $B$ is:
\begin{equation}
\mathcal{L}_{\mathrm{cross}}^{A \rightarrow B} = - \frac{1}{n} \sum_{i=1}^n \frac{\sum_{j=1}^m[\mathbf{M}_{\mathrm{cross}}^{A \rightarrow B}]_{ij} \cdot \ell_{ij}^{A \rightarrow B}}{\sum_{j=1}^m [\mathbf{M}_{\mathrm{cross}}^{A \rightarrow B}]_{ij} + \epsilon},
\end{equation}
\begin{equation}
\ell_{ij}^{A \rightarrow B} = \frac{(z_i^{A})^{\top} \hat{z}_j^{B}}{\tau} - \log \sum_{q=1}^{|\mathcal{Q}^B|} \exp \left(\frac{(z_i^{A})^{\top} \hat{z}_q^{B}}{\tau}\right),
\end{equation}
where $\hat{z}_j^{B}$ and $\hat{z}_q^{B}$ denote the features in memory bank $\mathcal{Q}^B$, and $\tau$ is the temperature parameter. The cross-domain loss from domain  B to domain  A  $\mathcal{L}_{\mathrm{cross}}^{B \rightarrow A}$ is computed symmetrically. Consequently, the total loss for cross-domain alignment is given by:
\begin{equation}
    \mathcal{L}_{\mathrm{cross}} = \mathcal{L}_{\mathrm{cross}}^{A \rightarrow B} + \mathcal{L}_{\mathrm{cross}}^{B \rightarrow A}.
\end{equation}
The progressive loss in the second stage is
defined as:
\begin{equation}
    \mathcal{L}_{\text{PA2}} = \mathcal{L}_{\text{in}} +  \lambda \mathcal{L}_{\text{cross}}.
\label{eq:PA2}
\end{equation}
After training, the outputs of $f_{\theta}$ are used for cross-domain retrieval.

\begin{table*}[ht]
\caption{Comparison of retrieval performance (P@50, P@100, P@200) across various domain on PACS. The best results are in \textbf{bold}.}
\label{tab:pacs}
\centering
\resizebox{0.9\linewidth}{!}{%
\begin{tabular}{l|ccc|ccc|ccc|ccc}
\toprule
\multirow{2}{*}{Method} & \multicolumn{3}{c|}{Art$\rightarrow$Cartoon} & \multicolumn{3}{c|}{Cartoon$\rightarrow$Art} & \multicolumn{3}{c|}{Art$\rightarrow$Photo} & \multicolumn{3}{c}{Photo$\rightarrow$Art} \\
& P@50 & P@100 & P@200 & P@50 & P@100 & P@200 & P@50 & P@100 & P@200 & P@50 & P@100 & P@200  \\
\midrule
ID & 26.32 & 25.45 & 24.66 & 27.13 & 25.93 & 24.43 & 19.17 & 18.78 & 18.48 & 19.65 & 19.62 & 18.81 \\
ProtoNCE & 28.45 & 25.38 & 22.63 & 27.63 & 24.69 & 21.96 & 28.14 & 25.35 & 24.33 & 30.50 & 26.90 & 23.53 \\
CDS & 45.84 & 40.96 & 35.93 & 38.01 & 36.09 & 32.00 & 41.01 & 37.30 & 33.27 & 84.32 & 79.60 & 65.91  \\
PCS & 48.98 & 44.38 & 39.20 & 42.00 & 39.55 & 35.13 & 45.43 & 41.82 & 37.56 & 84.70 & 79.95 & 66.75  \\
DD & 59.85 & 54.63 & 47.62 & 48.49 & 46.60 & 41.78 & 65.38 & 62.46 & 56.36 & 84.93 & 82.02 & 68.25  \\
CoDA & 53.73 & 48.99 & 42.81 & 45.49 & 42.98 & 38.07 & 58.23 & 55.04 & 50.18 & 84.93 & 79.90 & 67.53\\
ProtoOT & 69.97 & 65.94 & 61.37 & 55.95 & 55.91 & 55.08 & 84.23 & 83.99 & \textbf{79.88} & 86.08 & 85.50 & \textbf{83.89} \\
\textbf{DUDE} & \textbf{88.21} & \textbf{84.03} & \textbf{77.61} & \textbf{84.76} & \textbf{82.51} & \textbf{77.06} & \textbf{87.13} & \textbf{85.44} & 76.93 & \textbf{93.30} & \textbf{89.83} & 81.73  \\

\midrule
\multirow{2}{*}{} & \multicolumn{3}{c|}{Art$\rightarrow$Sketch} & \multicolumn{3}{c|}{Sketch$\rightarrow$Art} & \multicolumn{3}{c|}{Cartoon$\rightarrow$Photo} & \multicolumn{3}{c}{Photo$\rightarrow$Cartoon}\\
& P@50 & P@100 & P@200 & P@50 & P@100 & P@200 & P@50 & P@100 & P@200 & P@50 & P@100 & P@200  \\
\midrule
ID & 14.68 & 14.92 & 15.22 & 13.54 & 13.53 & 13.46 & 12.79 & 12.85 &13.56 & 15.16& 15.18& 15.37 \\
ProtoNCE & 33.73 & 29.72 & 25.34 & 25.94 & 23.40 & 24.35 & 36.23 & 30.75 & 27.05  & 40.62 & 34.31 & 28.04 \\
CDS & 19.06 & 18.73 & 19.99 & 18.69 & 17.44 & 16.09 & 24.66 & 24.52 & 23.63 & 45.89 & 41.39 & 34.24 \\
PCS & 20.60 & 20.15 & 21.02 & 18.95 & 17.79 & 16.25 & 25.68 & 25.60 & 24.76 & 47.21 & 42.52 & 36.91 \\
DD & 26.54 & 25.72 & 26.25 & 28.68 & 26.04 & 22.78 & 40.91 & 39.74 & 36.57 & 55.19 & 48.83 & 40.27  \\
CoDA & 23.26 & 22.68 & 23.14 & 25.87 & 22.95 & 20.01 & 36.27 & 35.74 & 33.63 & 55.57 & 49.08 & 40.39 \\
ProtoOT  & 23.51 & 24.06 & 26.65 & 30.41 & 29.86 & 27.16 & 42.04 & 42.71 & 43.18 & 64.74 & 60.25 & 54.58 \\
\textbf{DUDE} & \textbf{69.07} & \textbf{62.88} & \textbf{58.54} & \textbf{54.56} & \textbf{51.70} & \textbf{46.21} & \textbf{59.24} & \textbf{57.36} & \textbf{52.19} & \textbf{82.13} & \textbf{76.52} & \textbf{68.40}\\

\midrule
\multirow{2}{*}{} 
 & \multicolumn{3}{c|}{Cartoon$\rightarrow$Sketch}& \multicolumn{3}{c|}{Sketch$\rightarrow$Cartoon} & \multicolumn{3}{c|}{Photo$\rightarrow$Sketch}
&\multicolumn{3}{c}{Sketch$\rightarrow$Photo}\ \\
& P@50 & P@100 & P@200 & P@50 & P@100 & P@200 & P@50 & P@100 & P@200 & P@50 & P@100 & P@200  \\
\midrule
ID & 21.65 & 19.54 & 18.67 & 17.10 & 16.42 & 16.55 & 12.48 & 12.39 & 12.42  & 12.96 & 11.21 & 10.57  \\
ProtoNCE & 45.61 & 39.77 & 32.44 & 34.45 & 30.07 & 27.50 & 38.23 & 32.77 &27.08 & 25.34 & 21.96 & 22.58 \\
CDS & 34.09 & 31.51 & 30.27 & 22.56 & 21.08 & 20.54 & 18.84 & 18.77 & 20.77 & 12.54 & 10.57 & 9.79\\
PCS & 42.58 & 39.87 & 38.79 & 25.37 & 23.87 & 23.21 & 22.34 & 21.78 & 23.45 & 13.15 & 11.11 & 10.47\\
DD & 44.63 & 42.03 & 40.76 & 44.83 & 40.52 & 35.66 & 25.31 & 24.66 & 26.29 & 25.82 & 22.90 & 19.87\\
CoDA & 40.16 & 37.82 & 37.01 & 32.04 & 28.91 & 27.81 & 22.67 & 21.36 & 22.16 & 14.29 & 13.61 & 12.84 \\
ProtoOT & 50.02 & 47.10 & 46.07  & 53.54  & 50.90 & 44.71 & 27.74 & 24.31 & 19.63 & 14.87 & 14.79 & 14.35 \\
\textbf{DUDE} & \textbf{72.06} & \textbf{69.17} & \textbf{65.78}& \textbf{61.99} & \textbf{58.06} & \textbf{54.41} & \textbf{53.56} & \textbf{50.41} & \textbf{48.63} & \textbf{47.11} & \textbf{41.98} & \textbf{33.94} \\
\bottomrule
\end{tabular}
}
\end{table*}

\begin{table*}[ht]
\caption{Comparison of retrieval performance (P@1, P@5, P@15) across various domains on Office-Home. The best results are in \textbf{bold}.}
\label{tab:office_home_big}
\centering
\resizebox{0.9\linewidth}{!}{%
\begin{tabular}{l|ccc|ccc|ccc|ccc}
\toprule
\multirow{2}{*}{Method} & \multicolumn{3}{c|}{Art$\rightarrow$Real} & \multicolumn{3}{c|}{Real$\rightarrow$Art} & \multicolumn{3}{c|}{Art$\rightarrow$Product} & \multicolumn{3}{c}{Product$\rightarrow$Art} \\
& P@1 & P@5 & P@15 & P@1 & P@5 & P@15 & P@1 & P@5 & P@15 & P@1 & P@5 & P@15 \\
\midrule
ID & 35.89 & 33.13 & 29.60 & 39.89 & 34.42 & 27.65 & 25.88 & 24.91 & 22.49 & 32.17 & 25.94 & 20.23 \\
ProtoNCE & 40.50 & 36.39 & 34.00 & 44.53 & 39.26 & 32.99 & 29.54 & 27.89 & 25.75 & 35.73 & 30.61 & 24.55  \\
CDS & 45.08 & 41.15 & 38.73 & 44.71 & 40.75 & 35.53 & 32.76 & 31.47 & 28.90 & 35.75 & 32.48 & 26.82  \\
PCS & 41.70 & 38.51 & 36.22 & 44.96 & 39.88 & 33.99 & 33.29 & 31.50 & 29.53  & 39.24 & 34.77 & 28.77  \\
DD & 45.12 & 42.33 & 40.06 & 47.95 & 43.68 & 38.38 & 35.39 & 34.67 & 32.61 & 42.51 & 37.94 & 31.41  \\
CoDA & 44.77 & 40.99 & 36.64 & 44.88 & 37.54 & 38.12 & 34.52 & 33.96 & 31.06 & 40.98 & 32.24 & 30.54 \\
ProtoOT & {47.38} & {45.49} & {43.52} & {50.61} & {46.64} & {41.51} & {38.11} & {36.50} & {35.10} & {46.47} & {41.63} & {34.47}  \\
\textbf{DUDE} & \textbf{59.41} & \textbf{57.59} & \textbf{55.45} & \textbf{63.27} & \textbf{58.44} & \textbf{52.20}  & \textbf{46.39} & \textbf{46.55} & \textbf{45.98} & \textbf{57.55} &\textbf{53.93} & \textbf{47.31} \\

\midrule
\multirow{2}{*}{} & \multicolumn{3}{c|}{Clipart$\rightarrow$Real} & \multicolumn{3}{c|}{Real$\rightarrow$Clipart} & \multicolumn{3}{c|}{Product$\rightarrow$Real} & \multicolumn{3}{c}{Real$\rightarrow$Product}\\
& P@1 & P@5 & P@15 & P@1 & P@5 & P@15 & P@1 & P@5 & P@15 & P@1 & P@5 & P@15 \\
\midrule
ID & 29.48 & 26.48 & 23.25  & 35.51 & 32.17 & 27.96 & 50.73 & 45.03 & 39.05 & 45.12 & 41.46 & 38.01 \\
ProtoNCE& 25.25 & 22.66 & 20.83 & 41.15 & 37.66 & 31.95 & 53.84 & 48.25 & 42.21 & 47.74 & 44.85 & 41.21 \\
CDS & 32.51 & 30.30 & 27.80 & 38.88 & 36.48 & 33.16 & 54.00 & 50.07 & 45.60 & 49.39 & 47.27 & 43.98 \\
PCS & 29.07 & 26.06 & 24.00 & 40.60 & 38.11 & 34.06 & 56.45 & 50.78 & 45.37 & 49.90 & 47.11 & 43.73 \\
DD & 33.31 & 30.57 & 28.14 & 44.66 & 41.47 & 37.41 & 57.42 & 52.69 & 47.90 & 51.71 & 48.48 & 44.95  \\
CoDA & 30.12 & 27.10 & 24.02 & 43.65 & 35.21 & 29.06 & 57.37 & 50.98 & 42.82 & 55.23 & 49.18 & 44.36\\
ProtoOT  & {36.75} & {33.58} & {31.33} & {48.93} & {45.93} & {41.59} & {64.01} & {59.22} & {54.50} & {54.85} & {53.49} & {51.37} \\
\textbf{DUDE} & \textbf{43.11} & \textbf{41.59} & \textbf{41.03} & \textbf{57.10} & \textbf{55.57} & \textbf{52.57} & \textbf{74.97} & \textbf{72.77} & \textbf{68.51} & \textbf{68.60} & \textbf{67.19} & \textbf{64.83} \\

\midrule
\multirow{2}{*}{} 
 & \multicolumn{3}{c|}{Product$\rightarrow$Clipart}& \multicolumn{3}{c|}{Clipart$\rightarrow$Product} & \multicolumn{3}{c|}{Art$\rightarrow$Clipart}
&\multicolumn{3}{c}{Clipart$\rightarrow$Art}\ \\
& P@1 & P@5 & P@15 & P@1 & P@5 & P@15 & P@1 & P@5 & P@15 & P@1 & P@5 & P@15\\
\midrule
ID & 31.52 & 28.55 & 24.15 & 24.01 & 22.42 & 20.60 & 26.78 & 24.79 & 21.64 & 21.17 & 17.86 & 14.71 \\
ProtoNCE & 36.13 & 33.99 & 28.24 & 21.17 & 20.63 & 20.47 & 28.97 & 26.15 & 22.98 & 21.33 & 17.40 & 14.46\\
CDS & 37.69 & 34.99 & 30.42 & 27.24 & 26.46 & 24.86 & 25.59 & 23.77 & 22.41 & 22.41 & 20.34 & 17.34\\
PCS & 39.51 & 37.51 & 32.81 & 26.39 & 25.86 & 24.92 & 31.23 & 28.74 & 26.11 & 24.51 & 21.27 & 17.54\\
DD & 42.26 & 37.42 & 33.74 & 27.79 & 27.26 & 25.97 & 32.67 & 30.79 & 28.70 & 27.26 & 23.94 & 20.53\\
CoDA & 47.21 & 35.43 & 28.33 & 27.10 & 24.77 & 24.00 & 36.69 & 32.19 & 26.37 & 25.64 & 21.17 & 21.37 \\
ProtoOT & {44.76} & {42.64} & {38.83} & {29.92} & {30.15} & {29.29} & {35.39} & {34.57} & {32.05} & {28.96} & {25.58} & {22.21} \\
\textbf{DUDE} & \textbf{51.81} & \textbf{51.79} & \textbf{48.47} & \textbf{34.61} & \textbf{34.77} & \textbf{34.27} & \textbf{40.79} & \textbf{40.56} & \textbf{38.93} & \textbf{30.81} & \textbf{29.42} & \textbf{26.98}\\
\bottomrule
\end{tabular}
}
\end{table*}

\section{Experiments}
In this section, we first introduce the datasets, implementation details, evaluation metrics, and baseline methods. Then we evaluate the performance of our proposed DUDE against existing baselines on three datasets. Following this, we conduct ablation studies, parameter analysis, and visualization to further validate the effectiveness of DUDE.

\subsection{Datasets}
To evaluate the effectiveness of the proposed DUDE method, we conduct experiments on three widely used datasets in cross-domain retrieval, including PACS \cite{pacs}, Office-Home \cite{office}, and DomainNet \cite{domainnet}. Specifically, PACS consists of four different domains (Photo, Art Painting, Cartoon, Sketch) with seven categories. Office-Home comprises four domains (Art, Clipart, Product, Real) encompassing 65 categories. DomainNet \cite{domainnet} consists of six domains (Clipart, Infograph, Painting, Quickdraw, Real, and Sketch). Note that the Quickdraw domain is excluded due to its limited semantic. Consistent with prior studies \cite{dd,coda,protoot}, we utilize seven categories that contain over 200 images in DomainNet.

\subsection{Implementation Details}
In line with all baselines, a ResNet-50 \cite{resnet} network pre-trained on ImageNet is utilized to initialize the feature extractor $f_\theta$.
We utilize the parameters of $f_\theta$ to update $f_m$ following the MoCo strategy \cite{moco}. We train our model for 100 epochs using the Adam optimizer with a learning rate of $2.5 \times 10^{-4}$ and a batch size of 64. For fair comparisons, the temperature parameter $\tau$ is set as $0.2$ across all datasets. Without loss of generality, we apply Stable Diffusion v1.5~\cite{sd} in the object disentanglement module, which offers a good balance between generation speed and quality. 
In the first 50 epochs, we warm up the feature extractor $f_\theta$ using only the object disentanglement loss $\mathcal{L}_{\text{OD}}$. Afterward, we initialize $f_m$ with the parameters of the warmed-up $f_\theta$, and continue training $f_\theta$ using $\mathcal{L}_{\text{PA1}}$ for the next 30 epochs. Finally, we perform cross-domain alignment using $\mathcal{L}_{\text{PA2}}$ for the remaining 20 epochs.

\subsection{Evaluation Metrics}
Following previous studies, we evaluate the retrieval performance on OfficeHome using precision at the top 1, 5, and 15 retrieved images (denoted as P@1/5/15), and report precision at the top 50, 100, and 200 retrieved images for both PACS and DomainNet (denoted as P@50/100/200). Consistent with prior works, our task focuses on category-level cross-domain retrieval, where a retrieved image is considered correct if it belongs to the same class as the query.

\begin{table*}[ht]
\caption{Comparison of retrieval performance (P@50, P@100, P@200) across various domains on DomainNet. The best results are in \textbf{bold}.}
\label{tab:domainnet_big}
\centering
\resizebox{0.95\linewidth}{!}{%
\begin{tabular}{l|ccc|ccc|ccc|ccc}
\toprule
\multirow{2}{*}{Method} & \multicolumn{3}{c|}{Clipart$\rightarrow$Sketch} & \multicolumn{3}{c|}{Sketch$\rightarrow$Clipart} & \multicolumn{3}{c|}{Infograph$\rightarrow$Real} & \multicolumn{3}{c}{Real$\rightarrow$Infograph}  \\
& P@50 & P@100 & P@200 & P@50 & P@100 & P@200 & P@50 & P@100 & P@200 & P@50 & P@100 & P@200 \\
\midrule
ID & 49.46 & 46.09 & 40.44 & 54.38 & 47.12 & 37.73 & 28.27 & 27.44 & 26.33 & 39.98 &31.77 &24.84 \\
ProtoNCE & 46.85 & 42.67 & 36.35 & 54.52 & 45.04 & 35.06 & 28.41 & 28.53 & 28.50 & 57.01 &41.84 &30.33\\
CDS & 45.84 & 42.37 & 37.16 & 59.13 & 48.83 & 37.40 & 28.51 & 27.92 & 27.48 & 56.69& 39.76 &26.38 \\
PCS & 51.01 & 46.87 & 40.19 & 59.70 & 50.67 & 39.38 & 30.56 & 30.27 & 29.68 & 55.42 &42.13 &30.76 \\
DD & 56.31 & 52.74 & 47.38 & 63.07 & 57.26 & 48.17 & 35.52 & 35.24 & 34.35 & 57.74 &46.69 &35.47 \\
CoDA & 44.56 & 35.86 & 35.14 & 49.00 & 38.61 & 38.49 & 27.12 & 27.21 & 26.43 & 36.98 & 30.44 & 30.02 \\ 
ProtoOT & {70.46} & {69.41} & {67.41} & {82.79} & {78.68} & {71.31} & {40.65} & {40.35} & {40.05} & 77.02 & 67.33& 49.41\\
\textbf{DUDE} &\textbf{93.07}&  \textbf{92.25}&  \textbf{90.30}  & \textbf{95.21}& \textbf{94.56}&  \textbf{91.89} &\textbf{55.48} & \textbf{54.87} & \textbf{53.28} & \textbf{78.49} & \textbf{70.60} & \textbf{57.86}\\

\midrule
\multirow{2}{*}{} & \multicolumn{3}{c|}{Infograph$\rightarrow$Sketch} & \multicolumn{3}{c|}{Sketch$\rightarrow$Infograph} & \multicolumn{3}{c|}{Painting$\rightarrow$Clipart} & \multicolumn{3}{c}{Clipart$\rightarrow$Painting}\\
& P@50 & P@100 & P@200 & P@50 & P@100 & P@200 & P@50 & P@100 & P@200 & P@50 & P@100 & P@200   \\
\midrule
ID & 30.35 & 29.04 & 26.55  & 42.20 & 34.94 & 27.52 & 64.67 & 54.41 & 40.07 & 42.37 & 39.61 & 35.56 \\
ProtoNCE & 28.24 & 26.79 & 24.23 & 39.83 & 31.99 & 24.77 & 55.44 & 43.74 &  32.59 & 39.13 &  35.87 &  32.07 \\
CDS & 30.55 & 29.51 & 27.00 & 46.27 & 36.11 & 27.33 & 63.15 &  47.30 &  32.93 & 37.75 & 35.18 &  32.76  \\
PCS & 30.27 & 28.36 & 25.35 & 42.58 & 34.09 & 25.91 & 63.47 &  53.21 & 41.68 & 48.83 & 46.21 & 42.10 \\
DD & 31.29 & 29.33 & 26.54 & 43.66 & 36.14 & 28.12 & 66.42 & 56.84 & 46.72 & 52.58 & 50.10 & 46.11\\
CoDA & 24.94 & 22.48 & 22.42 & 27.65 & 23.85 & 23.48 & 57.30 & 44.25 & 44.15 & 46.51 & 42.38 & 41.22\\
ProtoOT & {37.16} & {36.30} & {34.42} & {63.59} & {53.30} & {38.75} & {90.21} & {87.38} & {77.14} & {71.13} & {71.15} & {70.50} \\
\textbf{DUDE} &\textbf{70.28}&  \textbf{68.48} & \textbf{66.40}  & \textbf{88.20} &  \textbf{85.24} & \textbf{76.01} &\textbf{93.23} & \textbf{89.99} &\textbf{83.10}& \textbf{87.01} & \textbf{84.59} & \textbf{80.36} \\
\bottomrule
\end{tabular}
}
\end{table*}

\begin{table*}[ht]
\caption{Average retrieval performance on PACS, Office-Home, and DomainNet.}
\label{tab:avg}
\centering
\resizebox{0.9\linewidth}{!}{%
\begin{tabular}{llcccccccc|c}
\toprule
{} & {Average} & {ID} &{ProtoNCE} & {CDS} & {PCS} &{DD} &{CoDA} &{ProtoOT} & \textbf{DUDE}  &{Improvement}\\
\midrule
\multirow{3}{*}{PACS}&P50&17.72 & 38.91 &33.79  & 36.42  & 45.87 & 41.04 & 50.26 & \textbf{71.09} & \textbf{+20.83} \\
&P@100&17.15 & 28.75 & 31.50 & 34.03 & 43.01 &38.25 & 48.78 & \textbf{67.49} & \textbf{+18.71} \\
&P@200& 16.85 & 25.56 & 28.54 & 31.12 & 38.54 & 34.63& 46.38 & \textbf{61.78} & \textbf{+15.40} \\
\midrule
\multirow{3}{*}{Office-Home}&P@1&33.18 &35.49 &37.17 &38.07 &40.67 & 40.68 &43.85&\textbf{52.36}&\textbf{+8.51}\\
&P@5&29.76 &32.15 &34.63 &35.01& 37.60 & 35.06 &41.29&\textbf{50.84}&\textbf{+9.55}\\
&P@15&25.78& 28.30 &31.30 &31.42 &34.15 & 31.39 &37.98&\textbf{48.04}&\textbf{+10.06}\\
\midrule
\multirow{3}{*}{DomainNet}&P@50&43.96&43.67&45.98&47.73&50.82 & 39.25 &66.63&\textbf{82.62}&\textbf{+15.99}\\
&P@100&38.80&37.05&38.37&41.47&45.54& 33.13 &62.98&\textbf{80.07}&\textbf{+17.09}\\
&P@200&32.37&30.48&31.05&34.38&39.10& 32.67&56.12&\textbf{74.90}&\textbf{+18.78}\\
\bottomrule
\end{tabular}
}
\end{table*}

\begin{table*}[ht]
\caption{Comparison of retrieval performance on the disentangled object features using different prompts or the same prompt on DomainNet.}
\label{tab:prompt_all1}
\centering
\resizebox{\linewidth}{!}{%
\begin{tabular}{l|ccc|ccc|ccc|ccc|ccc}
\toprule
\multirow{2}{*}{Prompt} & \multicolumn{3}{c|}{Clipart$\rightarrow$Sketch} & \multicolumn{3}{c|}{Sketch$\rightarrow$Clipart} & \multicolumn{3}{c|}{Infograph$\rightarrow$Real} & \multicolumn{3}{c|}{Real$\rightarrow$Infograph} & \multicolumn{3}{c}{Average} \\
& P@50 & P@100 & P@200 & P@50 & P@100 & P@200 & P@50 & P@100 & P@200  & P@50 & P@100 & P@200 & P@50 & P@100 & P@200\\
\midrule
varying & 75.76& 71.86& 65.37 &82.71 & 80.34& 74.08& 36.17& 36.40& 35.66 &75.21 & 65.17& 48.54 & 67.46 & 63.44 & 55.91\\
fixed & 66.42 & 63.45 & 58.44 & 76.70 & 74.35 & 68.34 & 34.87 & 34.46 &34.51 & 71.36 & 62.18 & 46.94 & 62.33 & 58.61 & 52.06\\

\midrule
\multirow{2}{*}{Prompt}  & \multicolumn{3}{c|}{Infograph$\rightarrow$Sketch} & \multicolumn{3}{c|}{Sketch$\rightarrow$Infograph} & \multicolumn{3}{c|}{Painting$\rightarrow$Clipart} & \multicolumn{3}{c|}{Clipart$\rightarrow$Painting}& \multicolumn{3}{c}{Average}\\
& P@50 & P@100 & P@200 & P@50 & P@100 & P@200 & P@50 & P@100 & P@200 & P@50 & P@100 & P@200 & P@50 & P@100 & P@200 \\
\midrule
varying & 42.78 &40.48 & 37.29& 64.52&56.82 &43.91 & 91.01 & 89.00 &82.11 &73.27 & 70.90& 65.71 & 67.89 & 64.30 & 57.25\\
fixed & 38.19 &36.53 & 34.13& 57.48 & 49.94 & 39.47 & 90.65 & 87.99 & 81.42 & 70.91 & 69.88 & 67.89 & 64.31 & 61.08 & 55.73\\
\bottomrule
\end{tabular}
}
\end{table*}

\begin{table*}[ht]
\caption{Ablation Study on our loss components. The best performance results are in \textbf{bold}.}
\label{tab:loss}
\centering
\resizebox{\linewidth}{!}{%
\begin{tabular}{ccc|ccc|ccc|ccc|>{\hskip -4pt}c
>{\hskip -10pt}c
>{\hskip -8pt}c
}
\toprule
\multirow{2}{*}{$\mathcal{L}_{\text{OD}}$} & \multirow{2}{*}{$\mathcal{L}_{\text{PA1}}$} & \multirow{2}{*}{$\mathcal{L}_{\text{PA2}}$} & \multicolumn{3}{c|}{PACS}
& \multicolumn{3}{c|}{Office-Home}&
\multicolumn{3}{c|}{DomainNet} & \multicolumn{3}{c}{Average} \\
 & & & P@50 & P@100 & P@200  & P@1 & P@5 & P@15 & P@50 & P@100 & P@200 & P@50(1) & P@100(5) & P@200(15)\\
\midrule
\multicolumn{3}{c|}{ResNet50} & 32.28 & 29.83 & 26.89  & 32.62 & 29.09 & 25.02 & 40.49 & 34.18 & 28.10 & 35.13 & 31.03 & 26.66\\
\midrule
\checkmark &  &  &  46.17 &
43.20 & 39.39 & 39.17 & 34.56 &
30.33 & 65.90 & 62.59 & 56.08&50.41&46.78&41.93 \\
 & \checkmark &  & 41.19 & 37.55 & 32.06 & 43.28 & 39.99 & 36.23  &  52.78 & 45.82 & 37.40&45.75&41.12&35.23\\
 &  & \checkmark & 47.67 & 44.08 & 39.02 & 35.88 & 32.50 & 28.76  & 53.49 & 48.21 & 40.92&45.68&41.60&36.23\\
\checkmark & \checkmark &   & 70.97 & 67.71 & 60.19 & 49.86 & 48.53 & 45.73 & 76.56 & 73.89 & 67.96&65.80&63.38&57.96 \\
{\checkmark} & & {\checkmark} & 50.23 & 47.85
& 44.32 & 44.83 & 42.13 & 38.62 & 79.16 & 77.06 & 71.45&58.07&55.68&51.46 \\
& \checkmark & \checkmark & 43.06 
& 39.34 & 33.76 & 44.52 & 41.59 &
37.37 & 57.89 & 51.67 & 42.88&48.49&44.20&38.00\\
\checkmark & \checkmark & \checkmark &  \textbf{71.09} & \textbf{67.79} & \textbf{61.78} & \textbf{52.36} & \textbf{50.84} & \textbf{48.04} & \textbf{82.62}
& \textbf{80.07} &\textbf{74.90}&\textbf{68.69}&\textbf{66.23}&\textbf{61.57}\\
\bottomrule
\end{tabular}
}
\end{table*}

\subsection{Baselines}
To evaluate the effectiveness of our DUDE, we compare it against the following baselines:  
\textit{i)} {ID} \cite{id}, which learns instance-level representations by distinguishing individual instances;
\textit{ii)} {ProtoNCE} \cite{protonce}, which enhances clustering performance by incorporating prototypes to capture semantic structure;
\textit{iii)} {CDS} \cite{cds}, which considers both intra-domain and cross-domain discriminative relations among instances to address domain adaptation; 
\textit{iv)} {PCS} \cite{pcs}, which uses K-means to generate prototypes and aligns domains using similarity vectors in a self-supervised manner;
\textit{v)} {DD} \cite{dd}, which introduces a distance-of-distance loss to quantify and minimize domain discrepancy;
\textit{vi)} {CoDA} \cite{coda}, which employs in-domain self-matching supervision and cross-domain classifier alignment to learn domain-invariant and discriminative features;
and \textit{vii)} {ProtoOT} \cite{protoot}, which unifies intra-domain and cross-domain representation learning through a prototype-based optimal transport framework to improve retrieval performance. For all baselines, we adopt the settings reported in their respective papers.

\subsection{Comparison with State-of-the-Art Methods}
Tabs.~\ref{tab:pacs}-\ref{tab:domainnet_big} provide detailed retrieval performance on PACS, Office-Home and DomainNet, respectively.  As summarized in Tab.~\ref{tab:avg}, DUDE achieves substantial performance improvements across all three datasets. On Office-Home, DUDE outperforms the leading baseline ProtoOT~\cite{protoot} with average gains of $8.50\%$, $9.55\%$, and $10.06\%$ in P@1, P@5, and P@15, respectively. On DomainNet, DUDE achieves average gains of $15.99\%$, $17.08\%$, and $18.78\%$ in P@50, P@100, and P@200, respectively. The improvements are even more pronounced on PACS, where DUDE achieves average gains of $20.83\%$, $18.71\%$, and $15.40\%$ in P@50, P@100, and P@200, respectively. 
The results demonstrate that the proposed method significantly enhances the performance of unsupervised cross-domain image retrieval, thanks to the disentangle-then-align strategy that narrows the domain gaps.

\subsection{Ablation Study}
To understand the impact of each component in the proposed DUDE, we conduct ablation studies on the disentanglement prompt and loss components.

\subsubsection{Disentanglement Prompt}
Recall that in our strategy to extract object features by disentangling them from domain-specific styles, we construct prompts in the format: ``a $\{domain\}$ of a $[z]$", where $\{domain\}$ specifies the style in domain (\textit{e.g.}, sketch, photo), and $[z]$ represents the extracted object features. The specific prompt templates used for the three datasets are summarized in Tab.~\ref{tab:text_prompt_set}.

To investigate whether the disentanglement indeed arises from the design of our varying $\{domain\}$ term, we compare retrieval performance based on $[z]$ learned under two prompt variants: \textit{i)} varying prompts in the form ``a $\{domain\}$ of a $[z]$", and \textit{ii)} a fixed prompt ``a photo of a $[z]$" applied across all domains. The results for DomainNet are shown in Tab.~\ref{tab:prompt_all1}. We observe that varying prompts consistently outperform the fixed prompt, particularly in domains involving highly different styles. For instance, when giving varying $\{domain\}$ terms, retrieval from Clipart to Sketch improves by 9.34$\%$ in P@50. This indicates that the image styles could be encapsulated by the varying domain-style token, enabling the model to learn cleaner and more disentangled object features compared to using a fixed prompt across domains.

\begin{figure}[t]
  \centering
   \includegraphics[width=\linewidth]{ 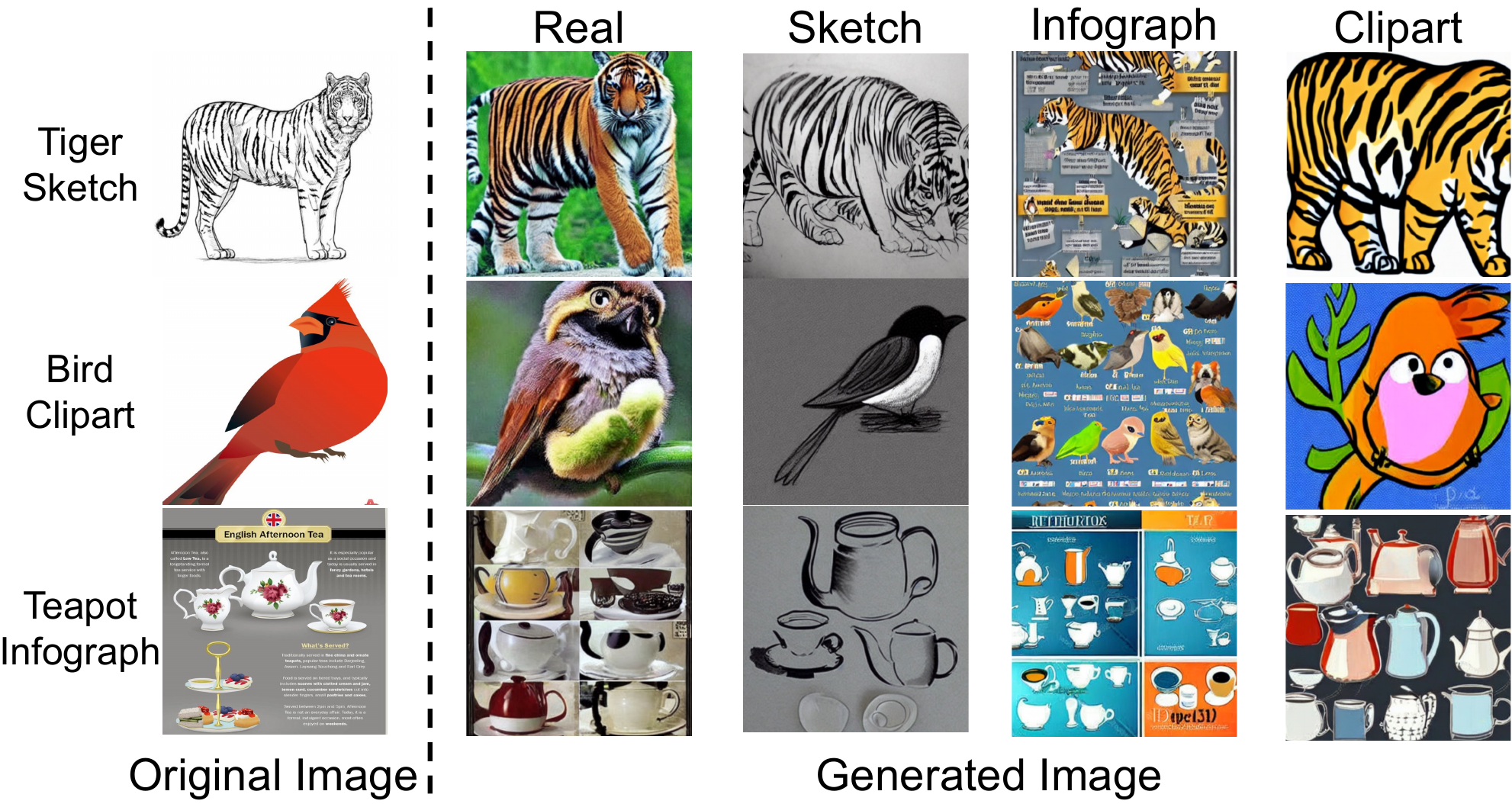}
   \caption{Cross-domain generation results using disentangled object features from one domain combined with other domain prompts. 
   }
   \label{fig:gene}
\end{figure}

To further visualize the effectiveness of our disentanglement strategy, we assess whether object features extracted from one domain can be recombined with other domain styles to generate cross-domain images with consistent object semantics. Specifically, given an image from a domain in DomainNet, we extract its object feature $[z]$ using $f_\theta$, and combine $[z]$ with four different $\{domain\}$ terms to form new prompts. These prompts are then used to generate images via SD, as illustrated in Fig.~\ref{fig:gene}. For example, starting with a sketch of a tiger, we generate a realistic photo, a new sketch, an infograph, and a clipart representation of the same tiger. Notably, the original sketch styles are absent in the generated images, confirming that our method successfully disentangles objects from domain-specific styles. 

\begin{figure*}[h]
  \centering
   \begin{subfigure}[b]{0.32\linewidth}
    \includegraphics[width=\linewidth]{ 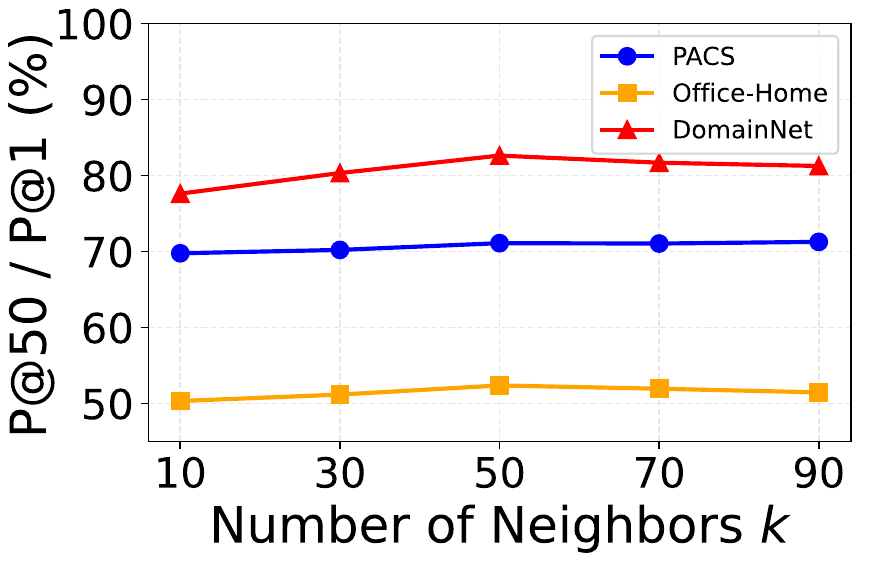}
    \caption{Analysis on neighbors $k$.}
    \label{fig:para1}
  \end{subfigure}
  \hfill
  \begin{subfigure}[b]{0.32\linewidth}
    \includegraphics[width=\linewidth]{ 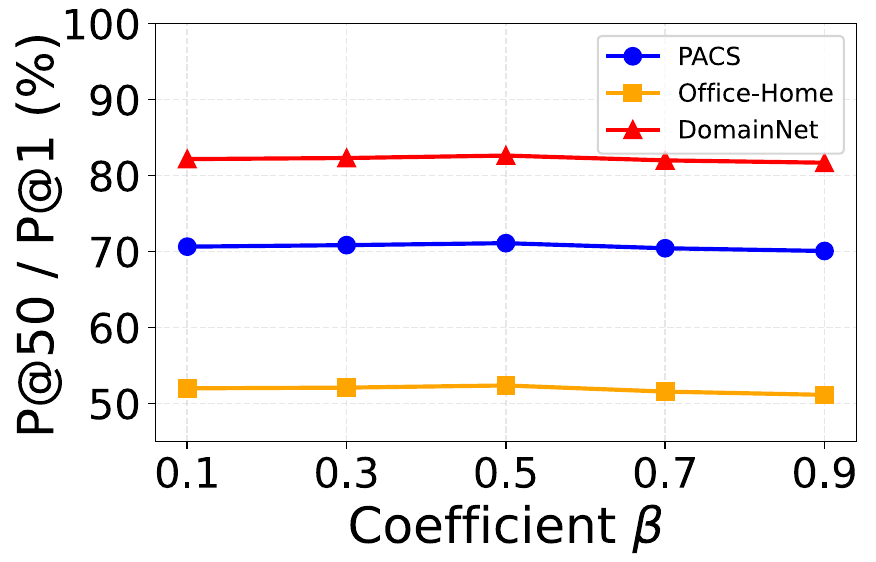}
    \caption{Analysis on coefficient $\beta$.}
    \label{fig:para2}
  \end{subfigure}
  \hfill
  \begin{subfigure}[b]{0.32\linewidth}
    \includegraphics[width=\linewidth]{ 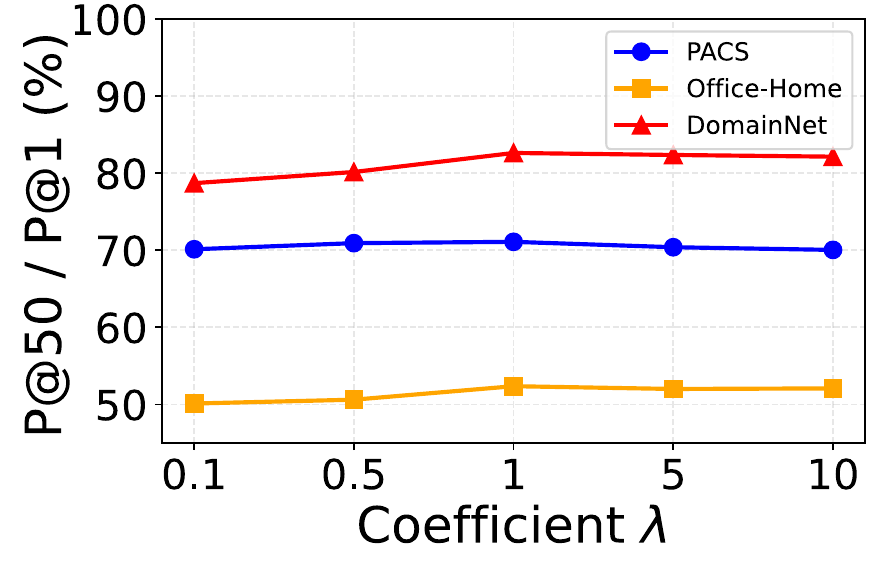}
    \caption{Analysis on coefficient $\lambda$.}
    \label{fig:para3}
  \end{subfigure}
   \caption{Parameter analysis on neighbors $k$, coefficient $\beta$ and coefficient $\lambda$.
   }
   \label{fig:para}
\end{figure*}

\subsubsection{Loss Components}
The results in Tab.~\ref{tab:loss} demonstrate the effectiveness of different loss components in our DUDE. From the results, one could
arrive at three conclusions:
\textit{i)} Comparing line 1 (a baseline using the pre-trained ResNet50 features) and line 2 shows that adopting $\mathcal{L}_{\text{OD}}$ alone leads to a substantial performance improvement, indicating that our disentanglement strategy helps the feature extractor learn clean, domain-invariant object features and significantly enhances feature discriminability;
\textit{ii)} Comparing Lines 3, 4, and 7, we first observe that the improvement from Line 3 to Line 7 underscores the benefit of our gradual transition from intra- to inter-domain alignment. However, compared to Line 7, Line 4 yields fluctuating results. For instance, a low P@1 of 35.88$\%$ on Office-Home versus a high P@50 of 47.67$\%$ on PACS. This disparity arises because the initial pre-trained ResNet50 features in Office-Home are not well aligned within domains, whereas PACS exhibits strong in-domain alignment. As a result, relying solely on $\mathcal{L}_{\text{PA2}}$ benefits PACS but leads to suboptimal optimization on Office-Home. These findings further underscore the necessity of our stable progressive alignment strategy;
\textit{iii)} Comparing lines 7 and 8, introducing $\mathcal{L}_{\text{OD}}$ leads to a notable performance boost, such as a 27$\%$ gain in P@50 on PACS. Similarly, comparing lines 2 and 8, jointly applying both $\mathcal{L}_{\text{PA1}}$ and $\mathcal{L}_{\text{PA2}}$ results in an improvement such as a 25$\%$ gain in P@50 on PACS. The optimal performance is observed when all three loss components are applied jointly, confirming that object disentanglement and progressive alignment are both indispensable.

\begin{figure*}[ht]
  \centering
   \includegraphics[width=0.8\linewidth]{ 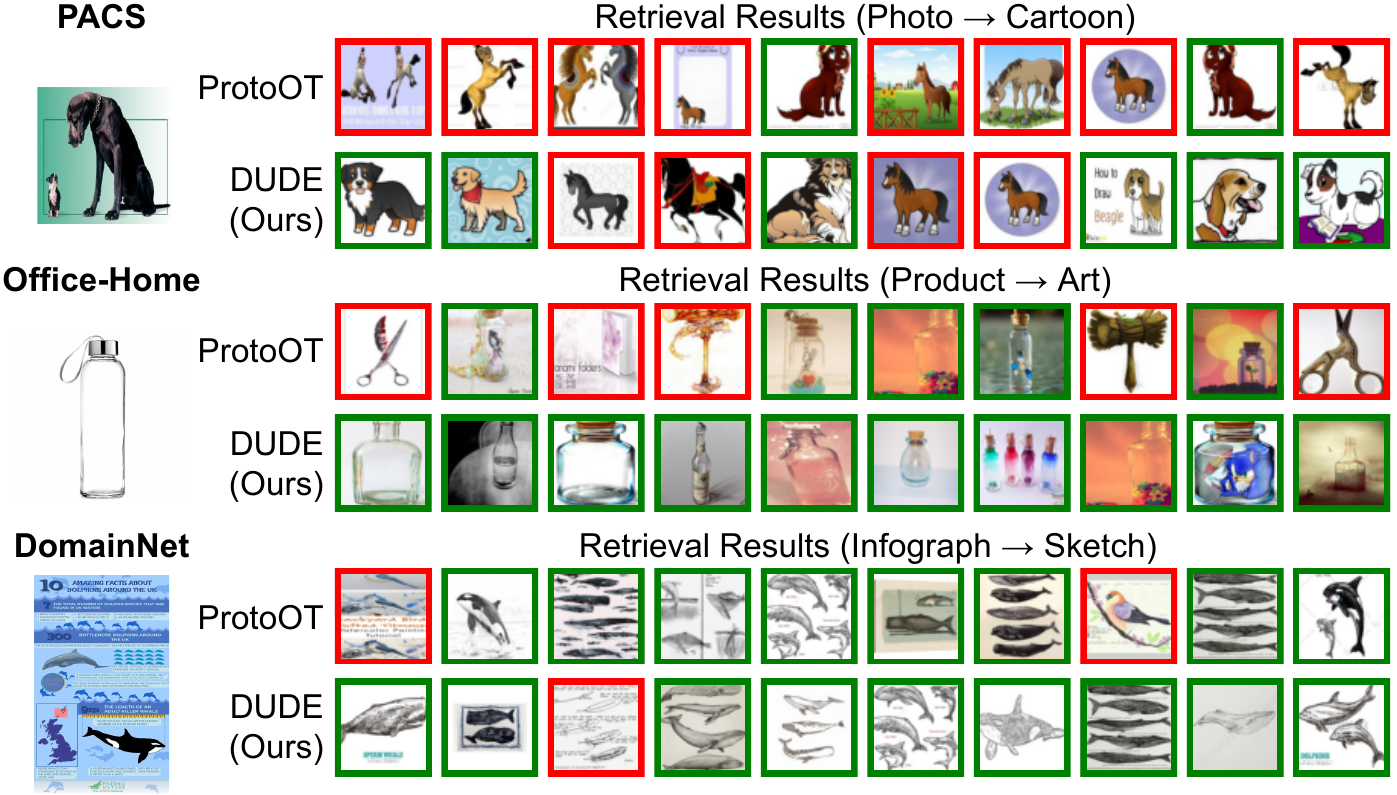}
   \caption{Top 10 unsupervised cross-domain image retrieval results using SOTA method ProtoOT and our DUDE. The retrievals are conducted on examples in three datasets.
The \textcolor{red}{red} borders indicate falsely retrieved results while the \textcolor{deepgreen}{green} borders indicate correctly retrieved results.
   }
   \label{fig:retrie}
\end{figure*}

\begin{figure*}[ht]
  \centering
   \includegraphics[width=0.9\linewidth]{ 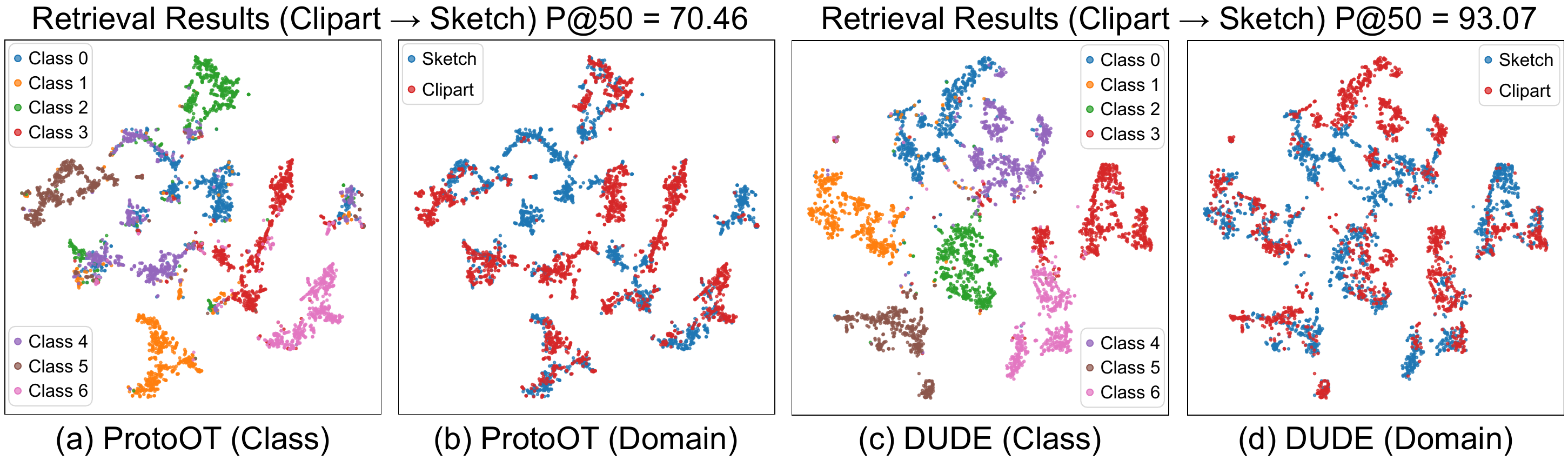}
  \caption{T-SNE visualization of learned features on Clipart-Sketch in DomainNet. (a) and (b) show the class- and domain-colored results of ProtoOT; (c) and (d) show the corresponding results of our DUDE.}
  \label{fig:tsne}
\end{figure*}

\subsection{Parameter Analysis}
To evaluate the robustness of DUDE, we conduct experiments by varying three key parameters: the number of neighbors $k$, the loss coefficient $\beta$ in $\mathcal{L}_{\text{PA1}}$ (Eq.~\ref{eq:PA1}), and the coefficient $\lambda$ in $\mathcal{L}_{\text{PA2}}$ (Eq.~\ref{eq:PA2}). The results are shown in Fig.~\ref{fig:para}.

Recall that DUDE aligns only those sample pairs that mutually identify each other as top-$k$ neighbors. In Fig.~\ref{fig:para1}, we evaluate the effect of varying $k$. As $k$ increases, the retrieval performance initially improves and then declines slightly. This can be attributed to two competing factors: on the one hand, larger $k$ introduces more potential true positive neighbors; on the other hand, overly large $k$ brings in noisy  neighbors. Nonetheless, our mutual top-$k$ neighbor strategy remains robust.
To evaluate the influence of $\beta$, we vary its value in $\{0.1, 0.3, 0.5, 0.7, 0.9\}$. As shown in Fig.~\ref{fig:para2}, DUDE demonstrates stable performance across all the values.
For $\lambda$, we evaluate values in $\{0.1, 0.5, 1, 5, 10\}$, as shown in Fig.~\ref{fig:para3}. DUDE performs well when $\lambda$ falls across this range.
Based on these observations, we set $k = 50$, $\beta = 0.5$, and $\lambda = 1$ for all experiments.

\subsection{Visualization}

To provide a more intuitive comparison between our DUDE and the most promising state-of-the-art method, ProtoOT \cite{protoot}, we present both retrieval visualizations on three datasets and t-SNE visualizations on the Clipart-Sketch pair from DomainNet.

\subsubsection{Retrieval Results Visualization}

We showcase the top-10 cross-domain retrieval results of our DUDE and ProtoOT. As illustrated in Fig.~\ref{fig:retrie}, the results are arranged top to bottom by dataset: Photo-Cartoon retrieval in PACS, Product-Art retrieval in Office-Home, and Infograph-Sketch retrieval in DomainNet. These examples demonstrate that DUDE consistently outperforms ProtoOT in most cases. For instance, in the query with the ``whale" infograph, ProtoOT fails to retrieve the correct category in the top-1 result, while DUDE retrieves correct instances within both top-1 and top-2. Even in more challenging cases, such as the third query of a ``dog" photo in PACS, DUDE retrieves 6 out of 10 correct matches, whereas ProtoOT only retrieves 2 out of 10.

\subsubsection{T-SNE Visualization}

To further illustrate the discriminative power of the learned embeddings, we visualize t-SNE projections for the Clipart-Sketch domains in DomainNet, colored by class and by domain. As shown in Fig.~\ref{fig:tsne}, (a) and (b) represent the class-wise and domain-wise t-SNE of ProtoOT, while (c) and (d) correspond to those of DUDE.
\textit{i)} Comparing (a) and (c), DUDE achieves more compact intra-class distributions and clearer inter-class separations, indicating improved class discriminability.
\textit{ii)} Comparing (b) and (d), in DUDE t-SNE, features exhibit better aggregation between different domains. This highlights DUDE’s effectiveness in reducing the domain gap through its disentanglement and progressive alignment strategies.

\section{Conclusion}
In this work, we present DUDE, a novel unsupervised cross-domain image retrieval (UCIR) method that explicitly addresses the domain gap through feature disentanglement. Unlike prior UCIR methods that extract and align features for the entire image, DUDE disentangles the domain-invariant object features from the domain-specific styles, by reversely mining the stylized image generation capability of diffusion models with the help of style-varying prompts. Further empowered by the proposed progressive alignment strategy, DUDE achieves reliable and accurate cross-domain retrieval results, significantly outperforming existing UCIR baselines. Looking ahead, DUDE not only offers a new perspective for tackling the UCIR task through feature disentanglement, but also opens up new application avenues for diffusion models beyond the image generation task.

\bigskip

\noindent \textbf{Author Contributions} All authors contributed to the study conception and design. The first draft of the manuscript was written by Ruohong Yang and all authors commented on previous versions of the manuscript. All
authors read and approved the final manuscript.

\bigskip

\noindent \textbf{Finding} No funding was received to assist with the preparation of this manuscript.

\bigskip

\noindent \textbf{Data Availability} The data that support the results and analysis of this study is publicly available in a repository. 
PACS dataset: \href{https://huggingface.co/datasets/flwrlabs/pacs}{https://huggingface.co/datasets/flwrlabs/pacs}. 
Office-Home dataset: \href{https://huggingface.co/datasets/flwrlabs/office-home}{https://huggingface.co/datasets/flwrlabs/office-home}.
DomainNet dataset: \href{https://ai.bu.edu/M3SDA/}{https://ai.bu.edu/M3SDA/DomainNet}.

\bigskip

\noindent 
\textbf{Compliance with Ethical Standards}

\noindent \textbf{Conflict of interest} The authors have no conflict of interest to declare
that are relevant to the content of this article.

\noindent \textbf{Ethics Approval} This article does not contain any studies with human participants performed by any of the authors.

\noindent \textbf{Consent to Participate} Informed consent was obtained from all individual participants included in the study.

\noindent \textbf{Consent for Publication} Consent for publication was obtained from the
participants.

\begin{appendices}
\section{Disentanglement Prompt Set}\label{secA1}
Our DUDE is built upon a text-to-image model, Stable Diffusion (SD). SD requires an image and a text condition as input. For each image in different domains, we construct a prompt with an object-semantic token as a text condition. The object-semantic token [$z$] represents the output of feature extractor $f_\theta$. Tab.~\ref{tab:text_prompt_set} summarizes the prompt set for three datasets.

\begin{table}[ht]
\caption{Disentanglement prompt set used in three datasets.}
\label{tab:text_prompt_set}
\centering
\begin{tabular}{lll} 
\toprule
\multicolumn{1}{l}{Dataset}& \multicolumn{1}{l}{Domain} & Prompt  \\ 
\midrule
\multirow{4}{*}{PACS} & Art Painting & a painting of a [$z$] \\
 & Cartoon & a cartoon of a [$z$]\\
 & Photo  & a photo of a [$z$]\\
 & Sketch & a sketch of a [$z$]\\ 
\midrule
\multirow{4}{*}{Office-Home} & Art & a painting of a [$z$] \\
& Clipart & a clipart~of a [$z$]        \\
 & Real   & a photo of a [$z$]          \\
& Product & a product photo of a [$z$]  \\
\midrule
\multicolumn{1}{l}{\multirow{5}{*}{DomainNet}} & Clipart& a clipart~of a [$z$] \\
\multicolumn{1}{l}{}& Sketch & a sketch of a [$z$] \\
\multicolumn{1}{l}{}& Painting& a painting of a [$z$]\\
\multicolumn{1}{l}{}& Infograph& an infograph of a [$z$]\\
\multicolumn{1}{l}{}& Real& a photo of a [$z$]\\ 
\bottomrule
\end{tabular}
\end{table}

\end{appendices}

\bibliographystyle{plain}
\bibliography{main}% common bib file

\begin{thebibliography}{10}

\bibitem{seg1}
Tomer Amit, Tal Shaharbany, Eliya Nachmani, and Lior Wolf.
\newblock Segdiff: Image segmentation with diffusion probabilistic models.
\newblock {\em arXiv preprint arXiv:2112.00390}, 2021.

\bibitem{cir6}
Chen Bao, Xudong Zhang, Jiazhou Chen, and Yongwei Miao.
\newblock Mmfl-net: multi-scale and multi-granularity feature learning for cross-domain fashion retrieval.
\newblock {\em Multimedia Tools and Applications}, 82(24):37905--37937, 2023.

\bibitem{pr1}
Junyi Chai, Reid Pryzant, Victor~Ye Dong, Konstantin Golobokov, Chenguang Zhu, and Yi~Liu.
\newblock Fast: Improving controllability for text generation with feedback aware self-training.
\newblock {\em arXiv preprint arXiv:2210.03167}, 2022.

\bibitem{user2}
Abhra Chaudhuri, Ayan~Kumar Bhunia, Yi-Zhe Song, and Anjan Dutta.
\newblock Data-free sketch-based image retrieval.
\newblock In {\em Proceedings of the IEEE/CVF Conference on computer vision and pattern recognition}, pages 12084--12093, 2023.

\bibitem{drepre2}
Prafulla Dhariwal and Alexander Nichol.
\newblock Diffusion models beat gans on image synthesis.
\newblock {\em Advances in neural information processing systems}, 34:8780--8794, 2021.

\bibitem{vae}
Carl Doersch.
\newblock Tutorial on variational autoencoders.
\newblock {\em arXiv preprint arXiv:1606.05908}, 2016.

\bibitem{cir5}
Bojana Gajic and Ramon Baldrich.
\newblock Cross-domain fashion image retrieval.
\newblock In {\em Proceedings of the IEEE conference on computer vision and pattern recognition workshops}, pages 1869--1871, 2018.

\bibitem{moco}
Kaiming He, Haoqi Fan, Yuxin Wu, Saining Xie, and Ross Girshick.
\newblock Momentum contrast for unsupervised visual representation learning.
\newblock In {\em Proceedings of the IEEE/CVF conference on computer vision and pattern recognition}, pages 9729--9738, 2020.

\bibitem{resnet}
Kaiming He, Xiangyu Zhang, Shaoqing Ren, and Jian Sun.
\newblock Deep residual learning for image recognition.
\newblock In {\em Proceedings of the IEEE conference on computer vision and pattern recognition}, pages 770--778, 2016.

\bibitem{beta}
Irina Higgins, Loic Matthey, Arka Pal, Christopher Burgess, Xavier Glorot, Matthew Botvinick, Shakir Mohamed, and Alexander Lerchner.
\newblock beta-vae: Learning basic visual concepts with a constrained variational framework.
\newblock In {\em International conference on learning representations}, 2017.

\bibitem{dd}
Conghui Hu and Gim~Hee Lee.
\newblock Feature representation learning for unsupervised cross-domain image retrieval.
\newblock In {\em European Conference on Computer Vision}, pages 529--544. Springer, 2022.

\bibitem{hu&small1}
Conghui Hu, Can Zhang, and Gim~Hee Lee.
\newblock Unsupervised feature representation learning for domain-generalized cross-domain image retrieval.
\newblock In {\em Proceedings of the IEEE/CVF International Conference on Computer Vision}, pages 11016--11025, 2023.

\bibitem{lora}
Edward~J Hu, Yelong Shen, Phillip Wallis, Zeyuan Allen-Zhu, Yuanzhi Li, Shean Wang, Lu~Wang, Weizhu Chen, et~al.
\newblock Lora: Low-rank adaptation of large language models.
\newblock {\em ICLR}, 1(2):3, 2022.

\bibitem{shop2}
Junshi Huang, Rogerio~S Feris, Qiang Chen, and Shuicheng Yan.
\newblock Cross-domain image retrieval with a dual attribute-aware ranking network.
\newblock In {\em Proceedings of the IEEE international conference on computer vision}, pages 1062--1070, 2015.

\bibitem{cir4}
Junshi Huang, Rogerio~S Feris, Qiang Chen, and Shuicheng Yan.
\newblock Cross-domain image retrieval with a dual attribute-aware ranking network.
\newblock In {\em Proceedings of the IEEE international conference on computer vision}, pages 1062--1070, 2015.

\bibitem{pr2}
Jaeseok Jeong, Junho Kim, Yunjey Choi, Gayoung Lee, and Youngjung Uh.
\newblock Visual style prompting with swapping self-attention, 2024.

\bibitem{cir2}
Xin Ji, Wei Wang, Meihui Zhang, and Yang Yang.
\newblock Cross-domain image retrieval with attention modeling.
\newblock In {\em Proceedings of the 25th ACM international conference on Multimedia}, pages 1654--1662, 2017.

\bibitem{ac1}
Takuhiro Kaneko, Kaoru Hiramatsu, and Kunio Kashino.
\newblock Generative attribute controller with conditional filtered generative adversarial networks.
\newblock In {\em Proceedings of the IEEE conference on computer vision and pattern recognition}, pages 6089--6098, 2017.

\bibitem{design1}
Tero Karras, Samuli Laine, Miika Aittala, Janne Hellsten, Jaakko Lehtinen, and Timo Aila.
\newblock Analyzing and improving the image quality of stylegan.
\newblock In {\em Proceedings of the IEEE/CVF conference on computer vision and pattern recognition}, pages 8110--8119, 2020.

\bibitem{ft1}
Nitish~Shirish Keskar, Bryan McCann, Lav~R Varshney, Caiming Xiong, and Richard Socher.
\newblock Ctrl: A conditional transformer language model for controllable generation.
\newblock {\em arXiv preprint arXiv:1909.05858}, 2019.

\bibitem{cds}
Donghyun Kim, Kuniaki Saito, Tae-Hyun Oh, Bryan~A Plummer, Stan Sclaroff, and Kate Saenko.
\newblock Cds: Cross-domain self-supervised pre-training.
\newblock In {\em Proceedings of the IEEE/CVF International Conference on Computer Vision}, pages 9123--9132, 2021.

\bibitem{law1}
Subhadeep Koley, Ayan~Kumar Bhunia, Aneeshan Sain, Pinaki~Nath Chowdhury, Tao Xiang, and Yi-Zhe Song.
\newblock Picture that sketch: Photorealistic image generation from abstract sketches.
\newblock In {\em Proceedings of the IEEE/CVF Conference on Computer Vision and Pattern Recognition (CVPR)}, pages 6850--6861, June 2023.

\bibitem{cir7}
Michal Kucer and Naila Murray.
\newblock A detect-then-retrieve model for multi-domain fashion item retrieval.
\newblock In {\em Proceedings of the IEEE/CVF conference on computer vision and pattern recognition workshops}, pages 0--0, 2019.

\bibitem{art2}
Minh-Ha Le and Niklas Carlsson.
\newblock Styleid: Identity disentanglement for anonymizing faces.
\newblock {\em arXiv preprint arXiv:2212.13791}, 2022.

\bibitem{protoot}
Bin Li, Ye~Shi, Qian Yu, and Jingya Wang.
\newblock Unsupervised cross-domain image retrieval via prototypical optimal transport.
\newblock In {\em Proceedings of the AAAI Conference on Artificial Intelligence}, volume~38, pages 3009--3017, 2024.

\bibitem{protonce}
Junnan Li, Pan Zhou, Caiming Xiong, and Steven~CH Hoi.
\newblock Prototypical contrastive learning of unsupervised representations.
\newblock {\em arXiv preprint arXiv:2005.04966}, 2020.

\bibitem{design2}
Xiang Li, Lei Meng, Lei Wu, Manyi Li, and Xiangxu Meng.
\newblock Dreamfont3d: personalized text-to-3d artistic font generation.
\newblock In {\em ACM SIGGRAPH 2024 Conference Papers}, pages 1--11, 2024.

\bibitem{small2}
Shaolei Liu, Siqi Yin, Linhao Qu, and Manning Wang.
\newblock Reducing domain gap in frequency and spatial domain for cross-modality domain adaptation on medical image segmentation.
\newblock In {\em Proceedings of the AAAI Conference on Artificial Intelligence}, volume~37, pages 1719--1727, 2023.

\bibitem{shop1}
Ziwei Liu, Ping Luo, Shi Qiu, Xiaogang Wang, and Xiaoou Tang.
\newblock Deepfashion: Powering robust clothes recognition and retrieval with rich annotations.
\newblock In {\em Proceedings of the IEEE conference on computer vision and pattern recognition}, pages 1096--1104, 2016.

\bibitem{ft2}
Haoming Lu, Hazarapet Tunanyan, Kai Wang, Shant Navasardyan, Zhangyang Wang, and Humphrey Shi.
\newblock Specialist diffusion: Plug-and-play sample-efficient fine-tuning of text-to-image diffusion models to learn any unseen style.
\newblock In {\em Proceedings of the IEEE/CVF Conference on Computer Vision and Pattern Recognition}, pages 14267--14276, 2023.

\bibitem{disen}
Michael~F Mathieu, Junbo~Jake Zhao, Junbo Zhao, Aditya Ramesh, Pablo Sprechmann, and Yann LeCun.
\newblock Disentangling factors of variation in deep representation using adversarial training.
\newblock {\em Advances in neural information processing systems}, 29, 2016.

\bibitem{small4}
Hyeonseob Nam, HyunJae Lee, Jongchan Park, Wonjun Yoon, and Donggeun Yoo.
\newblock Reducing domain gap by reducing style bias.
\newblock In {\em Proceedings of the IEEE/CVF conference on computer vision and pattern recognition}, pages 8690--8699, 2021.

\bibitem{infonce}
Aaron van~den Oord, Yazhe Li, and Oriol Vinyals.
\newblock Representation learning with contrastive predictive coding.
\newblock {\em arXiv preprint arXiv:1807.03748}, 2018.

\bibitem{law2}
Shuxin Ouyang, Timothy~M Hospedales, Yi-Zhe Song, and Xueming Li.
\newblock Forgetmenot: Memory-aware forensic facial sketch matching.
\newblock In {\em Proceedings of the IEEE Conference on Computer Vision and Pattern Recognition}, pages 5571--5579, 2016.

\bibitem{domainnet}
Xingchao Peng, Qinxun Bai, Xide Xia, Zijun Huang, Kate Saenko, and Bo~Wang.
\newblock Moment matching for multi-source domain adaptation.
\newblock In {\em Proceedings of the IEEE International Conference on Computer Vision}, pages 1406--1415, 2019.

\bibitem{class}
Mihir Prabhudesai, Tsung-Wei Ke, Alexander~Cong Li, Deepak Pathak, and Katerina Fragkiadaki.
\newblock Test-time adaptation of discriminative models via diffusion generative feedback.
\newblock In {\em Thirty-seventh Conference on Neural Information Processing Systems}, 2023.

\bibitem{seg2}
Aimon Rahman, Jeya Maria~Jose Valanarasu, Ilker Hacihaliloglu, and Vishal~M Patel.
\newblock Ambiguous medical image segmentation using diffusion models.
\newblock In {\em Proceedings of the IEEE/CVF conference on computer vision and pattern recognition}, pages 11536--11546, 2023.

\bibitem{sd}
Robin Rombach, Andreas Blattmann, Dominik Lorenz, Patrick Esser, and Bj{\"o}rn Ommer.
\newblock High-resolution image synthesis with latent diffusion models.
\newblock In {\em Proceedings of the IEEE/CVF conference on computer vision and pattern recognition}, pages 10684--10695, 2022.

\bibitem{user3}
Aneeshan Sain, Ayan~Kumar Bhunia, Pinaki~Nath Chowdhury, Subhadeep Koley, Tao Xiang, and Yi-Zhe Song.
\newblock Clip for all things zero-shot sketch-based image retrieval, fine-grained or not.
\newblock In {\em Proceedings of the IEEE/CVF conference on computer vision and pattern recognition}, pages 2765--2775, 2023.

\bibitem{ac2}
Huajie Shao, Shuochao Yao, Dachun Sun, Aston Zhang, Shengzhong Liu, Dongxin Liu, Jun Wang, and Tarek Abdelzaher.
\newblock Controlvae: Controllable variational autoencoder.
\newblock In {\em International conference on machine learning}, pages 8655--8664. PMLR, 2020.

\bibitem{user}
Arnold~WM Smeulders, Marcel Worring, Simone Santini, Amarnath Gupta, and Ramesh Jain.
\newblock Content-based image retrieval at the end of the early years.
\newblock {\em IEEE Transactions on pattern analysis and machine intelligence}, 22(12):1349--1380, 2000.

\bibitem{office}
Hemanth Venkateswara, Jose Eusebio, Shayok Chakraborty, and Sethuraman Panchanathan.
\newblock Deep hashing network for unsupervised domain adaptation.
\newblock In {\em Proceedings of the IEEE Conference on Computer Vision and Pattern Recognition}, pages 5018--5027, 2017.

\bibitem{cir1}
Haixin Wang, Hao Wu, Jinan Sun, Shikun Zhang, Chong Chen, Xian-Sheng Hua, and Xiao Luo.
\newblock Idea: An invariant perspective for efficient domain adaptive image retrieval.
\newblock {\em Advances in Neural Information Processing Systems}, 36:57256--57275, 2023.

\bibitem{ucir1}
Xu~Wang, Dezhong Peng, Ming Yan, and Peng Hu.
\newblock Correspondence-free domain alignment for unsupervised cross-domain image retrieval.
\newblock In {\em Proceedings of the AAAI Conference on Artificial Intelligence}, volume~37, pages 10200--10208, 2023.

\bibitem{coda}
Xu~Wang, Dezhong Peng, Ming Yan, and Peng Hu.
\newblock Correspondence-free domain alignment for unsupervised cross-domain image retrieval.
\newblock In {\em Proceedings of the AAAI Conference on Artificial Intelligence}, volume~37, pages 10200--10208, 2023.

\bibitem{style}
Zhouxia Wang, Xintao Wang, Liangbin Xie, Zhongang Qi, Ying Shan, Wenping Wang, and Ping Luo.
\newblock Styleadapter: A unified stylized image generation model.
\newblock {\em arXiv preprint arXiv:2309.01770}, 2023.

\bibitem{video2}
Jay~Zhangjie Wu, Yixiao Ge, Xintao Wang, Stan~Weixian Lei, Yuchao Gu, Yufei Shi, Wynne Hsu, Ying Shan, Xiaohu Qie, and Mike~Zheng Shou.
\newblock Tune-a-video: One-shot tuning of image diffusion models for text-to-video generation.
\newblock In {\em Proceedings of the IEEE/CVF International Conference on Computer Vision}, pages 7623--7633, 2023.

\bibitem{id}
Zhirong Wu, Yuanjun Xiong, Stella~X Yu, and Dahua Lin.
\newblock Unsupervised feature learning via non-parametric instance discrimination.
\newblock In {\em Proceedings of the IEEE conference on computer vision and pattern recognition}, pages 3733--3742, 2018.

\bibitem{pacs}
Jiaolong Xu, Liang Xiao, and Antonio~M L{\'o}pez.
\newblock Self-supervised domain adaptation for computer vision tasks.
\newblock {\em IEEE Access}, 7:156694--156706, 2019.

\bibitem{small3}
Qinwei Xu, Ruipeng Zhang, Ya~Zhang, Yanfeng Wang, and Qi~Tian.
\newblock A fourier-based framework for domain generalization.
\newblock In {\em Proceedings of the IEEE/CVF conference on computer vision and pattern recognition}, pages 14383--14392, 2021.

\bibitem{art1}
Wenju Xu, Chengjiang Long, and Yongwei Nie.
\newblock Learning dynamic style kernels for artistic style transfer.
\newblock In {\em Proceedings of the IEEE/CVF conference on computer vision and pattern recognition}, pages 10083--10092, 2023.

\bibitem{drepre}
Sihyun Yu, Sangkyung Kwak, Huiwon Jang, Jongheon Jeong, Jonathan Huang, Jinwoo Shin, and Saining Xie.
\newblock Representation alignment for generation: Training diffusion transformers is easier than you think.
\newblock {\em arXiv preprint arXiv:2410.06940}, 2024.

\bibitem{pcs}
Xiangyu Yue, Zangwei Zheng, Shanghang Zhang, Yang Gao, Trevor Darrell, Kurt Keutzer, and Alberto~Sangiovanni Vincentelli.
\newblock Prototypical cross-domain self-supervised learning for few-shot unsupervised domain adaptation.
\newblock In {\em Proceedings of the IEEE/CVF conference on computer vision and pattern recognition}, pages 13834--13844, 2021.

\bibitem{drepre3}
Junyi Zhang, Charles Herrmann, Junhwa Hur, Luisa Polania~Cabrera, Varun Jampani, Deqing Sun, and Ming-Hsuan Yang.
\newblock A tale of two features: Stable diffusion complements dino for zero-shot semantic correspondence.
\newblock {\em Advances in Neural Information Processing Systems}, 36:45533--45547, 2023.

\bibitem{video1}
Yabo Zhang, Yuxiang Wei, Dongsheng Jiang, Xiaopeng Zhang, Wangmeng Zuo, and Qi~Tian.
\newblock Controlvideo: Training-free controllable text-to-video generation.
\newblock {\em arXiv preprint arXiv:2305.13077}, 2023.

\bibitem{cir3}
Xiaoping Zhou, Xiangyu Han, Haoran Li, Jia Wang, and Xun Liang.
\newblock Cross-domain image retrieval: methods and applications.
\newblock {\em International Journal of Multimedia Information Retrieval}, 11(3):199--218, 2022.

\end{thebibliography}

\end{document}